# Learning Generative Models of Similarity Matrices

**Rómer Rosales**
Dept. of Elect. and Comp. Engineering
University of Toronto
romer@psi.toronto.edu

**Brendan Frey**
Dept. of Elect. and Comp. Engineering
University of Toronto
frey@psi.toronto.edu

## Abstract

Recently, spectral clustering (a.k.a. normalized graph cut) techniques have become popular for their potential ability at finding irregularly-shaped clusters in data. The input to these methods is a similarity measure between every pair of data points. If the clusters are well-separated, the eigenvectors of the similarity matrix can be used to identify the clusters, essentially by identifying groups of points that are related by transitive similarity relationships. However, these techniques fail when the clusters are noisy and not well-separated, or when the scale parameter that is used to map distances between points to similarities is not set correctly. Our approach to solving these problems is to introduce a generative probability model that explicitly models noise and can be trained in a maximum-likelihood fashion to estimate the scale parameter. Exact inference is computationally intractable, but we describe tractable, approximate techniques for inference and learning. Interestingly, it turns out that greedy inference and learning in one of our models with a fixed scale parameter is equivalent to spectral clustering. We examine several data sets, and demonstrate that our method finds better clusters compared with spectral clustering.

## 1 INTRODUCTION AND RELATED WORK

One way to analyze a set of training data is to measure the similarity between every pair of data points, and perform analysis on the resulting similarity matrix. Examples of similarity measures include the inverse of the Euclidean distance between two data vectors and the exponential of the negative distance. By throwing away the original data, this approach is unable to make use of the results of analysis to tease out valuable statistics in the input data that can further refine the analysis. However, the hope is is that the similarities are sufficient for performing data analysis.

The use of similarity matrices for data analysis has arisen in various fields, including cluster analysis (Ng, Jordan, and Weiss, 2002; Meila and Shi, 2001b), analysis of random walks (Chung, 1997; Meila and Shi, 2001b), dimensionality reduction (Tenenbaum, Silva, and Langford, 2000; Brand, 2003; Belkin and Niyogi, 2001), (also (Roweis and Saul, 2000)), segmentation in computer vision (Shi and Malik, 2000; Meila and Shi, 2001a), and link analysis (Kleinberg, 1998; Brin and Page, 1998). Similarity matrices are also called affinity matrices, adjacency matrices or transition matrices, depending on the context. In this paper, we will concentrate on clustering using spectral analysis, or 'spectral clustering' (SC), but the concepts and algorithms introduced are applicable to these other areas.

Before spectral clustering can be applied, the data points $\mathcal{X} = \{x_1, x_2, ..., x_N\}$ are mapped to an $N \times N$ affinity matrix, $\mathbf{L}$. For example, we may set $L_{ij} \propto e^{-d(x_i,x_j)/\gamma^2}$, where $d(x_i, x_j)$ is a non-negative distance measure, and $\gamma$ is a scale parameter that must, somehow, be chosen. The scaling parameter is implicitly related to the concept of neighborhood size. In spectral clustering, $\mathbf{L}$ is normalized.

Roughly speaking, the goal is to partition the data into $M$ clusters, such that within each cluster, each point is related to every other point by a chain of highly similar points in the same cluster. The variable $c_k \in \{1, ..., M\}$ represents the class label of the $k$th data point, and $\mathbf{C} = (c_1, ..., c_N)$ is the variable representing the class labels for all data points. If the data is well-separated, the eigenvectors of the similarity matrix will identify the different clusters. Once normalized, the similarity matrix can be viewed as the transition matrix of a stochastic process, and the eigenvectors will identify the different stationary distributions corresponding to the separate clusters.

In general, spectral clustering consists of finding the eigenvalue decomposition $\mathbf{VDV}^\top$ of the affinity matrix $\mathbf{L}$ [1], where the columns of $\mathbf{V}$ contain the eigenvectors. All but the first few eigenvectors are retained for analysis. The $N$ rows of $\mathbf{V}$ (which are associated to the $N$ data points) are then clustered using a simple *metric* method. Alternative definitions of SC can be found (Meila and Shi, 2001b; Shi and Malik, 2000; Kannan, Vempala, and Vetta, 2000). In (Ng, Jordan, and Weiss, 2002) however, an extra normalization step takes place, each row of $\mathbf{V}$ is normalized before the spatial clustering step. If the clusters are well-separated, the rows of $\mathbf{V}$ are orthogonal and identify the different clusters, so they can be clustered easily. The row-clustering method is secondary, and varies in the literature.

---

[1] The matrix $\mathbf{L}$ is assumed normalized. Some authors employ

Despite its extensive use, there is no clear understanding of the class of problems that would benefit from using spectral clustering (Ng, Jordan, and Weiss, 2002). One of our interests is to develop a generative modeling view of spectral clustering, with the aim of providing a maximum likelihood interpretation, which is useful for estimating the scale parameter and noise statistics. (Note that although (Meila and Shi, 2001b) analyzes the behavior of spectral clustering using stochastic processes, this approach does not define a generative model of the similarity matrix.)

In the rest of this paper we will introduce two families of probability models for $\mathbf{L}$ and $\mathbf{C}$ that we found of theoretical and practical interest. The first is based on latent feature vectors, where each data point has a corresponding low-dimensional feature vector that must be inferred. The second is based on a representation of the transitive similarity relationships described above, as a latent graph that must be inferred. The first allows us to give a generative model interpretation of spectral clustering and we show that greedy inference in our generative model gives a standard spectral clustering algorithm. The second allows us to generalize spectral clustering, and in particular enables us to find maximum likelihood estimates of the scale parameters, and account for noisy, overlapping clusters using a noise model.

## 2 LATENT FEATURE REPRESENTATIONS

This class of probabilistic models for $\mathbf{L}$ and $\mathbf{C}$ is based on the concept that there is an alternative (spatial) representation for each $x_i$ (*e.g.*, a global transformation) that could *expose* the cluster structure in the data set.

Formally, let us assume that there is a $t$-dimensional vector (feature) $\lambda_i \in \Re^t$ associated with each data point $x_i$. Denote $\Lambda$ the matrix whose $i$–th column contains the vector $\lambda_i$. In this class of models, we do not observe $\lambda_i$ but some function of groups of points from $\mathcal{X}$. For a given data set $\mathcal{X}$, let this function be given by $L_{ij}$. The goal is to find $M$ (class-dependent) probability distributions over $\lambda_i$ ($i = 1, ..., N$) to explain the observations given by $\mathbf{L}$ as

---

$\mathbf{L} \leftarrow S^{-1}\mathbf{L}$, others use $\mathbf{L} \leftarrow S^{-1/2}\mathbf{L}S^{-1/2}$, $s_{ii} = \sum_j L_{ij}$.

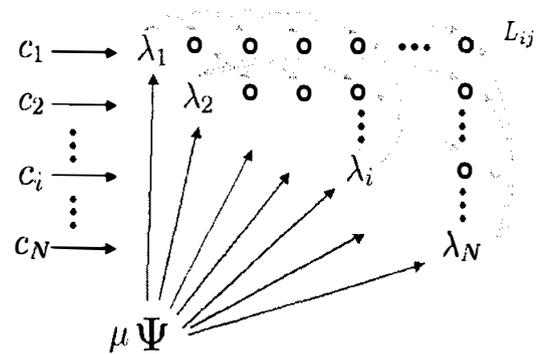

Figure 1: Latent feature representation probability model. Each circle represents one element $L_{ij}$ of the normalized matrix $\mathbf{L}$. In this figure, $\mathbf{L}$ is symmetric, thus only the elements for which $i < j$ need to be represented.

generated from one of the $M$ classes. Because of the nature of the hidden variables $\lambda_i$, we will call this class of models, latent feature representation models. In order to make this model explicit, let us consider the joint probability distribution over $\mathbf{C}, \Lambda, \mathbf{L}, \mu, \Psi$ associated to the graphical model in Fig. 1 given by:

$$p(\mathbf{C}, \Lambda, L, \mu, \Psi) = p(\mu)p(\Psi)\prod_{ij}p(L_{ij}|\lambda_i, \lambda_j)$$
$$\prod_i p(\lambda_i|c_i, \mu, \Psi)P(c_i), \quad (1)$$

with $\mu = (\mu_1, ..., \mu_M)$ and $\Psi = (\psi_1, ..., \psi_M)$ defining the class-dependent probability distributions over the hidden variable $\Lambda$. If $\mathbf{L}$ is symmetric we only consider terms with $i < j$. This model indicates that each $\lambda_i$ was generated by one of $M$ class-dependent conditional distributions with parameters $\mu_i, \psi_i$ and that each entry $L_{ij}$ in the $\mathbf{L}$ matrix was, in turn, a function of both $\lambda_i$ and $\lambda_j$. Both of these functions and the class-dependent conditional distributions are not specified so far. As will be seen next, this class of models provides a probabilistic interpretation to widely used clustering methods.

### 2.1 SPECTRAL CLUSTERING AS GREEDY INFERENCE IN A PROBABILISTIC MODEL

Despite wide use, the SC algorithm is lacking an interpretation in terms of uncertainty. The SC algorithm was not formulated in a probabilistic setting. Given the growing interest in SC, recently there has been several attempts to justify its use from other viewpoints (Ng, Jordan, and Weiss, 2002; Brand, 2003) and understand the class of problems where it is guaranteed to perform well. Here, we show how our latent feature model can provide a means for analysis by explicitly defining a probabilistic model for SC.

The following lemma says that with the appropriate choices for conditional distributions in our latent feature model,

greedy inference in this model is equivalent to the general form of the spectral clustering algorithm defined in Sec. 1 (*c.f.*, (Meila and Shi, 2001b; Shi and Malik, 2000; Kannan, Vempala, and Vetta, 2000)).

**Lemma 1** *The spectral clustering algorithm defined in Sec. 1 is equivalent to greedy probabilistic inference in the model with joint probability distribution defined in Eq. 1 with conditional distributions:*

$$p(L_{ij}|\lambda_i, \lambda_j) = \mathcal{N}(L_{ij}; \lambda_i^\top \lambda_j, \Sigma_\Lambda) \quad (2)$$
$$p(\lambda_i|c_i, \mu, \Psi) = \mathcal{N}(\lambda_i; \mu_{c_i}, \psi_{c_i}), \quad (3)$$

*and uniform priors for* **C**, $\mu$, *and* $\Psi$ *(in a bounded interval).*

First, let us make clear that **L** is observed and that it is normalized. We can quickly observe that greedy inference in the probabilistic model of Fig. 1 consist on (1) finding a MAP estimate of $\Lambda$ given **L** and (2) using this estimate of $\Lambda$ to estimate **C** and also $\mu, \Psi$.

After choosing a dimensionality $t$ for each $\lambda_i$, we can see that step 1 is equivalent to:

$$\arg\min_{\Lambda \in \Re^{N \times t}} \sum_{ij} (L_{ij} - \lambda_i^\top \lambda_j)^2 = \arg\min_{\hat{\mathbf{L}}} ||\mathbf{L} - \hat{\mathbf{L}}||_F, \quad (4)$$

where $||A||_F$ denotes the Frobenius norm of matrix $A$ and $\hat{\mathbf{L}}$ has rank $t$.

A well known linear algebra fact is that the best rank-$t$ approximation of $A$ with respect to the Frobenius norm is given by the eigenvalue decomposition of $A$. If the eigenvalue decomposition of **L** is given by $\mathbf{VDV}^\top$ (**V** and **D** are assumed ordered from largest to smallest eigenvalue), then the optimal matrix $\Lambda^*$ is equal to $\mathbf{D}^{1/2}\mathbf{V}^\top$, where only the $t$ largest eigenvalues and corresponding eigenvectors are considered. In other words, $\lambda_i^*$ is the vector formed by the scaled $i - th$ component (dimension) of the $t$ eigenvectors with largest eigenvalues. A seen before, a rank $t$ eigenvalue decomposition of the affinity matrix **L** is similarly the first step of spectral clustering algorithms.

In greedy inference, it is assumed that $\Lambda^*$ is the only probable value for the random variable $\Lambda$, thus step (2) consists on inferring **C**, $\mu$ and $\Psi$ from knowledge of $\Lambda^*$. If we only consider MAP estimation of $\mu$ and $\Psi$ we have the standard mixture model density estimation problem. This is a standard clustering technique, also equivalent to $k$-means if $\psi_k \to 0$. These are two out or many possible choices for this conditional distribution. SC similarly uses a spatial clustering algorithm to group the eigenvector rows found in the first step. Thus, both algorithms are equivalent up the choice of the final spatial grouping method. □

We have seen that inference in the latent feature representation model presented here is intimately related to the spectral clustering algorithm. This connection is particularly important since it allows us to use progress in probabilistic inference and apply it to the spectral clustering setting. For example, we can think of generalizations of spectral clustering by seeking new (approximate[2]) algorithms for inference in our probabilistic model (*e.g.*, algorithms rooted on approximate methods such as variational inference). Also, it allows us to generalize spectral clustering by replacing Eqs. 2,3 to more adequate, perhaps problem dependent, choices. Due to space limitations, we will not develop these extensions here. However, in the next section we introduce a related probabilistic model which incorporates a more general view of the clustering problem by using structured hidden representations.

## 3 LATENT GRAPH REPRESENTATIONS

The general view of clustering emphasized in the previous section serves as preamble for thinking about different hidden representation forms. Instead of looking for hidden feature-based representations, one can imagine extending this notion to less local and not *location* based representations. This idea is what we intend to develop in the rest of the paper. As apparent from the title, now the latent random variable represents a graph.

### 3.1 MOTIVATION

The use of spectral clustering methods has been encouraged because: (1) they are well defined approximations to graph partitioning methods (Chung, 1997), (2) there are simple (approximate) algorithms that obtain accurate solutions, and (3) success have been demonstrated in numerous data sets; even when there is no clear justification that the SC method is appropriate ((Ng, Jordan, and Weiss, 2002) is an exception) or that it has the desired clustering properties. However, SC does not seem to perform well on surprisingly simple datasets, like those shown in Fig. 2(c-f).

It is possible, however, to formally see why SC does not provide suitable answers for these apparently simple problems. From a random-walk perspective (Meila and Shi, 2001b; Chung, 1997), we can see that these two sets seem to (approximately) maximize the probability of staying in the same set given that a random walker starts his 'journey' at any point within the set [3] (with probability proportional to that of reaching the given starting point). As can be seen from Fig. 2(c-f), this does not seem to be the right goal or cost function. Is it possible to find a probabilistic model that is more suitable for these class of problems?

---

[2] Since the maximum clique size of the related moralized graph is $N$, inference complexity is exponential in $N$, the number of data points. This is in agreement with the original clustering problem complexity.

[3] We assumed that the random walker has probability to go to another data point proportional to the inverse distance between the points. Probability can be associated with any other measure that we decide it is useful.

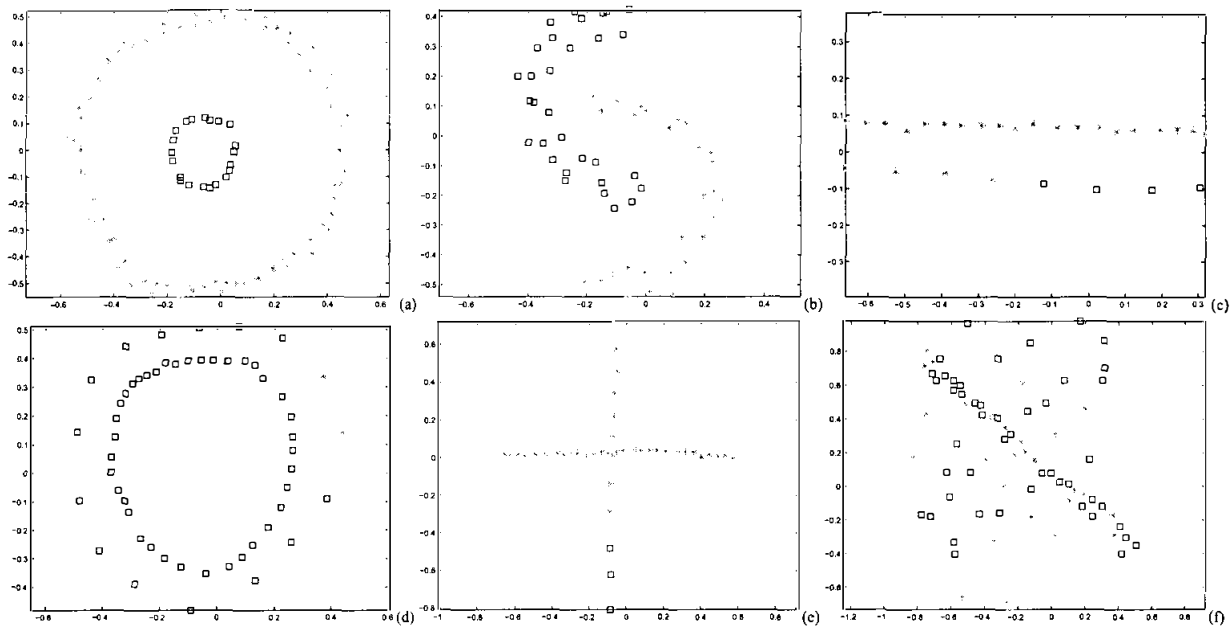

Figure 2: Spectral clustering (SC) examples. (a) and (b): Example problems where SC performs well. (c-f) Example problems where SC performance poorly. The clusters found are represented by a specific symbol per cluster (and color where available). A thorough search was conducted over $\Re$ to obtain the best parameter $\gamma$ as proposed in (Ng, Jordan, and Weiss, 2002).

## 3.2 SCALED DISTANCE / SIMILARITY MATRIX

Let us imagine that we do not really know the measure that we want to use in clustering the observed data points, or simply that there is no obvious choice for this measure. One simple way to represent this lack of knowledge is, for example, to assume that there is a unknown measure that is inherent to each class. One can also make the more general statement that the unknown measure varies even within classes. However, in this paper we will concentrate on the simpler notion of one underlying measure, sometimes referred to as *scale*, (possibly one for each class) with an appropriate uncertainty parameter.

Another notion that we wish to incorporate in a clustering paradigm is that not all entries in the affinity matrix should be equally important. For example, if we know that data point $i$ belongs to a cluster different than that of data point $j$, then the entry $L_{ij}$ should not *matter* as much as entry $L_{ik}$ with point $k$ in the same cluster as $i$. If data points $i$ and $j$ belong to different clusters, then the fact that they are or they are not close to each other should not matter much for the clustering algorithm. A similar argument can be applied to the relationship between points in the same class. However, in order to use these potentially desirable properties, we need to know the clusters and measure in advance (and obviously this is not the case). Thus, the sense of how much an entry $L_{ij}$ matters should also be inferred by the clustering algorithm. This idea is not incorporated in the underlying SC cost function where all entries matter according to the chosen measure, and the measure is fixed and common to all the elements in the matrix.

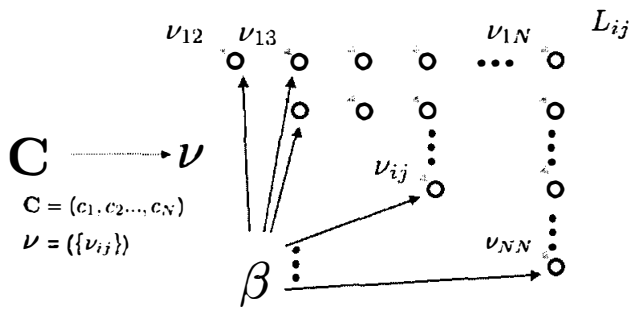

Figure 3: Latent graph representation probability model. Each circle represents one element $L_{ij}$ of $\mathbf{L}$. Each variable $L_{ij}$ has one variable $\nu_{ij}$ pointing to it (arrows drawn for some entries only). In its general form $P(\nu|\mathbf{C})$ does not need to factorize, however several factorizations are proposed in this paper. To ease representation two directed graphs are shown (they should be interpreted as one).

Both of these ideas have been captured in the probability model represented by the graph in Fig. 3. The joint distribution implied by this graph is given by:

$$p(\mathbf{C}, \mathbf{L}, \beta, \nu) = p(\mathbf{L}|\beta, \nu)P(\nu|\mathbf{C})p(\beta)P(\mathbf{C}) \quad (5)$$
$$= \prod_{ij} p(L_{ij}|\nu_{ij}, \beta)P(\nu|\mathbf{C})K_C K_\beta, (6)$$

where $L_{ij}$ represents, as usual, each entry in $\mathbf{L}$, $\beta = (\beta_1, ..., \beta_M)$ represents the basic *scale* parameter for each class concept of similarity (not always a distance metric), $\nu$ is the random variable representing latent neighborhoods among data points (in particular $\nu_{ij}$ indicates if data point $i$ and $j$ are neighbors), and $c_i$ represents the class label for data point $i$. In this model we define $L_{ij} = |x_i - x_j|_2$ (thus, $L$ should be called a distance matrix in this case), although this is not critical for the model proposed here, i.e., the exponential form defined in Sec. 1 could also be used.

We will use the following representation for $\nu$: $\nu_{ij} = c \Rightarrow c_i = c_j = c$. Representing the class directly in $\nu$ has the advantage that it avoids the dependence on $C$ of the conditional distribution of $L_{ij}$. This dependence would have created extra loops in our graph, and further complicated the exposition of inference methods in the next section [4].

With the above representation, it is simple to define $p(L|\nu, \beta)$. In this paper we will explore two definitions:

$$p(L_{ij}|\nu_{ij}, \beta) = \mathcal{N}(L_{ij}; \beta_{\nu_{ij}}, \sigma^2_{\nu_{ij}}) \quad (7)$$

$$p(L_{ij}|\nu_{ij}, \beta) = \beta_{\nu_{ij}} e^{-\beta_{\nu_{ij}} L_{ij}}, \quad (8)$$

i.e., a Gaussian and an exponential conditional distributions. Note that $\beta_0$ and $\sigma_0^2$ (where applicable) define a background distribution. This distribution is associated to entries in $L$ that come from points in different clusters.

$p(L|\nu, \beta)$ effectively defines a measure in $\Re$, and is connected with the concept of similarity between two points. For example, in the case of Eq. 7, two points $x_i$ and $x_j$ in class $k$ are more similar, the closer $L_{ij}$ is to $\beta_k$. In this case, our model will prefer configurations where the distances between neighbors in a class are similar. In the case of Eq. 8, $\beta_k$ determines how fast the similarity value decreases with respect to $L_{ij}$ (a notion more closely related to spectral clustering and its scale parameter $\gamma$).

The conditional distribution $P(\nu|C)$ plays the key role in determining what are the *admissible neighborhoods* allowed by the model. These neighborhoods can be viewed in terms of *admissible graphs* and thus the term *latent graph representations*. Note that this is a different graph from the Bayes network graph used in describing the joint probability distributions. Thus, $P(\nu|C)$ effectively defines a family of admissible graphs over the set of data points. As an example, one can enforce the constrain that every data point has $K$ neighbors. This can be accomplished by the following choice:

$$P(\{\nu_{ij}\}_{j=1..N}|C) = \begin{cases} T & \text{if } \sum_j I(c_i = c_j = \nu_{ij}) = K \\ 0 & \text{otherwise,} \end{cases} \quad (9)$$

with $K \in \{1..N\}$, $I$ the indicator function, and $T$ defined so that this is a valid uniform probability distribution (no symmetry between neighbors is assumed). This implies the following factorization: $P(\nu|C) = \prod_i P(\{\nu_{ij}\}_{j=1..N}|C)$.

In our experimental results, we perform some tests using this prior. However, the drawback of this choice is that for different problems, a different $K$ may be required. A more suitable form of admissible graphs are those that only enforce that points in the same class must be connected somehow (i.e., directly or through links traversing other same-class points). We thus arrive to the following choice:

$$P(\nu|C) = \begin{cases} T & \text{if } \forall c; G^{(c)}(V^{(c)}, E^{(c)}) \text{ is connected} \\ 0 & \text{otherwise,} \end{cases} \quad (10)$$

with the graph $G^{(c)}$ for class $c \in \{1, ..., M\}$ defined in terms of nodes $V^{(c)} = \{i|c_i = c\}$, edges $E^{(c)} = \{e_{ij}|\nu_{ij} = c\}$, and constant $T$ as above. This simply says that the links connecting points in the same class must form a connected graph. In the next section, we will see that it is possible to formalize an approximate, efficient inference algorithm for finding $C$.

### 3.3 INFERENCE IN THE SCALED MODEL

Due to the complex interactions between $L$ and $C$, it is difficult to infer $C$ exactly. More specifically, $L$ depends on all the class labels $C = (c_1, ..., c_N)$. Therefore, there is no straightforward factorization of the joint distribution, and moreover, this function is not differentiable. We can note that inference would take at least exponential time $N$ by looking at the maximum clique size in the graph of Fig. 3 (*n.b.*, hidden in our representation of $C$, there is a fully connected graph).

In the following, we will derive a simple approximate inference algorithm for this model. Our goal is obtaining a MAP estimate for $C$, $\beta$, and $\nu$ after observing the similarity matrix $L$. We will start with our definition of the log joint probability distribution. Assuming the Gaussian likelihood in Eq. 7, we have:

$$\log p(C, L, \beta, \nu) = -\frac{1}{2}[\sum_{ij} \frac{(L_{ij} - \beta_{\nu_{ij}})^2}{\sigma^2_{\nu_{ij}}} + \log \sigma^2_{\nu_{ij}}]$$
$$+ \log P(\nu|C) + K_C + K_\beta \quad (11)$$

Let us first assume we can compute $P(\nu, C|\beta, L)$ somehow. Based on this distribution we can use the well known EM algorithm and derive a EM-like update on $\beta$ and $\sigma^2$:

$$\beta_k \leftarrow \frac{\sum_{ij} \tilde{P}(\nu_{ij} = k) L_{ij}}{\sum_{ij} \tilde{P}(\nu_{ij} = k)} \quad (12)$$

$$\sigma_k^2 \leftarrow \frac{\sum_{ij} \tilde{P}(\nu_{ij} = k)(L_{ij} - \beta_k)^2}{\sum_{ij} \tilde{P}(\nu_{ij} = k)} \quad (13)$$

However, it remains unknown how to compute $p(\nu, C, \beta|L)$, a difficult task, since the space and time complexity for storing and computing the full distribution is exponential in $N$. However, we will see that finding the MAP estimate is perhaps surprisingly simple.

First, we can show the following for the special case when $\sigma^2$ and $\beta$ are the same for all classes:

---

[4] In either case, however, exact inference has similar exponential complexity.

**Theorem 1** *Let $\sigma_k^2 = \sigma^2$, $\beta_k = \beta$ ($k = 1,...,M$), then for some $\sigma_0$, $\arg\max_\nu P(\mathbf{C}, \nu|\mathbf{L}, \beta_{\nu_{ij}})$ is equivalent to finding the minimum spanning tree in a fully connected graph with weights $w_{ij} = \log p(L_{ij}|\nu_{ij}, \beta_{\nu_{ij}}) - \log p(L_{ij}|\nu_{ij} = 0, \beta_0)$.*

Proof: According to our definition, we have:

$$\log P(\mathbf{C}, \nu|\mathbf{L}, \beta_{\nu_{ij}}) \quad (14)$$

$$= \sum_{ij}(\log p(L_{ij}|\nu_{ij}, \beta_{\nu_{ij}}) - \log p(L_{ij}|\nu_{ij} = 0, \beta_0))$$

$$+ \sum_{ij} \log p(L_{ij}|\nu_{ij} = 0, \beta_0) + \log P(\nu|\mathbf{C})$$

$$= \sum_{ij} -f(L_{ij}, \nu_{ij}) + \log P(\nu|\mathbf{C}) + Const. \quad (16)$$

Note that if $\nu_{ij} = 0$, then the $(ij)$ term will not contribute to the sum. If $\nu_{ij} \neq 0$ then that term contributes, and since $\sigma_k^2 = \sigma^2$ and $\beta_k = \beta$, the contribution is the same for any value of $\nu_{ij}$. Thus, we want to find $\arg\min_{\mathbf{C}} \sum_{ij|\nu_{ij} \neq 0} f(L_{ij}, \nu_{ij}) - \log P(\nu|\mathbf{C})$ (after switching the signs). In words, we want the smaller sum of edges, with the condition that the resulting graph is connected.

If $f(a, b) \geq 0$ for $a \in \Re$, then clearly this problem is equivalent to finding the minimum spanning tree (MST) of a fully connected graph with (symmetric) weights $w_{ij} = f(L_{ij}, \nu_{ij})$ and with nodes indexed by $i$ and $j$.

We will see that this condition on $f(a, b)$ cannot be met in general, but it is possible to satisfy it for any subset of the domain $\Re$ that we are interested in.

In our problem the above condition can be written:

$$p(L_{ij}|\nu_{ij}, \beta_{\nu_{ij}}) \leq p(L_{ij}|\nu_{ij} = 0, \beta_0); \forall i, j \quad (17)$$

Clearly it is not possible to satisfy Eq. 17 for any value of the random variable in $\Re$ since each side of the inequality must be a probability distribution (except for the uninteresting case when both sides of the inequality are identical). However, it is possible to guarantee that for any specific problem with finite $L_{ij}$ (for any $i, j$), there exist $\sigma_0^2$ such that Eq. 17 is satisfied.

Using the definitions for $p(L_{ij}|\nu_{ij}, \beta_{\nu_{ij}})$ we have that Eq. 17 implies:

$$\alpha_{ij} \geq \frac{\log \sigma_0^2 - \log \sigma_{\nu_{ij}}^2}{\sigma_0^2 - \sigma_{\nu_{ij}}^2}(\sigma_0^2 \sigma_{\nu_{ij}}^2) = g(\sigma_0^2, \sigma_{\nu_{ij}}^2), \quad (18)$$

with $\alpha_{ij} = (L_{ij} - \beta_{\nu_{ij}})^2$. It can be shown that $g$ is a monotonic, strictly concave function of $\sigma_0^2 \in \Re_+ - \{\sigma_{\nu_{ij}}^2\}$ for any fixed $\sigma_{\nu_{ij}}^2$. Thus, the inverse $g^{-1}(\alpha_{ij}, \sigma_{\nu_{ij}}^2)$ exists for any particular $\sigma_{\nu_{ij}}^2$. This function is convex (and monotonically increasing) on $\alpha_{ij} \in \Re_+$. Therefore, for any $\alpha_{ij} > 0$ and $\sigma_{\nu_{ij}}^2$, it is possible to find $\sigma_0^2$ such that Eq. 17 is satisfied. The case $\alpha_{ij} = 0$ occurs if $L_{ij} = \beta$, however, this is a set of measure zero. Note that it suffices to find $\sigma_0^2$ for $\Gamma = \min_{ij} \alpha_{ij}$ since all $\alpha_{ij}$ will also satisfy the inequality if $\Gamma$ does.

We have so far assumed the conditional distribution in Eq. 7, the same steps can be used to prove the theorem for Eq. 8. In any case, Eq. 17 is satisfied and the MST algorithm is the correct MAP estimator for $\nu$. For completeness, it can also be shown that the labels are determined by splitting the spanning tree at the $M - 1$ most expensive edges. □

The above property is not applicable directly for inference, since we have $M$-different class-conditional distributions $p(L_{ij}|\nu_{ij} = k, \beta_0)$, $k = 1, ..., M$ (i.e., with different parameters). We have the following corollary which is valid for any (Gaussian) class-conditional distributions for $L_{ij}$.

**Corollary 1** *For a given assignment of $\mathbf{C}$ and any Gaussian conditional probability distribution $p(L_{ij}|\mathbf{C}, \nu_{ij} = c, \beta_c)$ for each class $c$, there exist $\sigma_0$ such that $\arg\max_\nu P(\mathbf{C}, \nu|\mathbf{L}, \beta)$ is equivalent to finding $M$-minimum spanning trees independently, one for each class, for a fully (in-class) connected graph with weights $w_{ij} = \log p(L_{ij}|\nu_{ij} = c, \beta_c) - \log p(L_{ij}|\nu_{ij} = 0, \beta_0)$.*

Proof: This just says that if we somehow knew the class assignments, then the most likely graph would consist of the union of all the MST for each class. This can be proven directly from the above theorem by observing that given $\mathbf{C} = (c_1, ..., c_N)$ then the rhs of Eq. 11 becomes:

$$\sum_{c=1}^{M}[\sum_{ij|\nu_{ij}=c} \log p(L_{ij}|\nu_{ij}, \beta_{\nu_{ij}}) + P_c(\nu|\mathbf{C})], \quad (19)$$

with $P_c(\nu|\mathbf{C})$ the prior for the graph associated to cluster $c$. As shown, this decomposes into $M$ separate sums that can be solved using the MST algorithm according to Theorem 1. Note that given $\mathbf{C}$, any $\nu_{ij}$ can only take two possible values, $\nu_{ij} = c_i = c_j$ or $\nu_{ij} = 0$. □

Using these results, we arrive to the following simple inference algorithm:

| Scaled Affinity Matrix Inference Algorithm |
|---|
| 1. For each $i = 1, ..., N$, initialize $c_i$ with a random cluster *i.e.*, a value in $\{1, ..., M\}$. |
| 2. For iters. 1,2,...., update $\mathbf{C}, \nu, \beta, \sigma^2$ as follows:<br>• Pick a random $i$ and find the best graph $\nu$ and class $c_i$ by solving $\arg\max_{\nu, c_i} P(\mathbf{C}, \nu|\mathbf{L}, \beta)$ using Corollary 1.<br>• Update $\beta_k$ and $\sigma_k^2$ using Eqs. 12-13 |

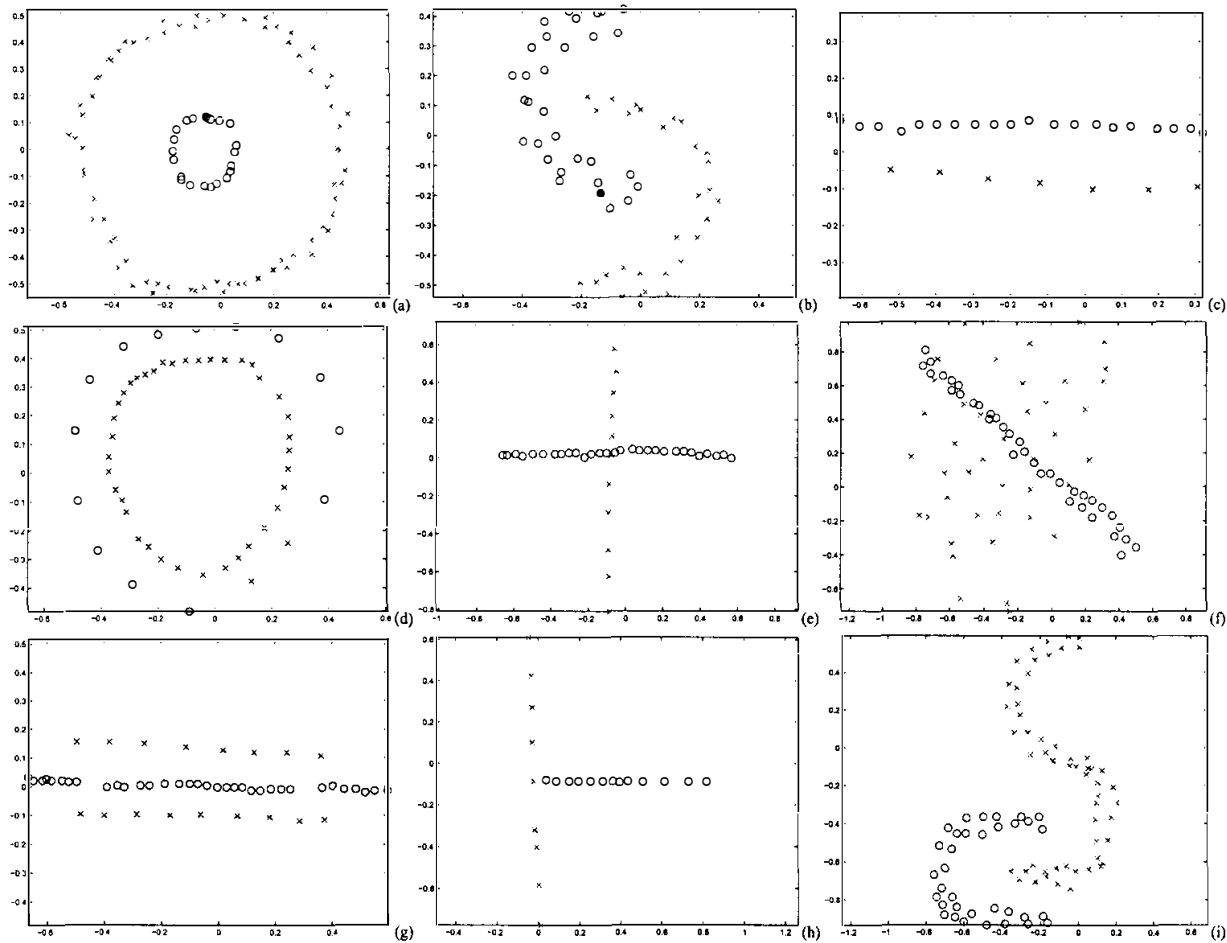

Figure 4: Clustering results obtained using our inference method. Performance is clearly better when compared to those in Fig. 2. Points in each cluster are represented by a different symbol (and color where available).

## 4 EXPERIMENTAL RESULTS

In order to experimentally test our learning/inference algorithm, we applied it to a sensible variety of clustering problems. In many of them, related algorithms perform poorly. Except when indicated, we used the connected graph prior specified in Eq. 10. Note that in this case, we do not require any parameter setting, unlike most previously presented methods which require parameter tuning (Shi and Malik, 2000; Meila and Shi, 2001b) or automatic (although exhaustive) parameter search (Ng, Jordan, and Weiss, 2002). In our experiments, we use $M = 2$ classes, however it is straightforward to extend the algorithm to more classes; each iteration is $O(MN \log N)$. The algorithm scales well for data points in higher dimensions, since its complexity does not depend on dimensionality. Results are shown in Fig. 4. We also show the most likely graph $\nu$ for one of the classes using dotted links. The results are excellent, even for problems where state-of-the-art methods do not perform well.

One of the most robust clustering algorithms currently is spectral clustering (SC). This was reasonably shown in (Ng, Jordan, and Weiss, 2002), where another version of SC (Meila and Shi, 2001b) and other methods such as K-means, and connected components were tested. Thus, we decided to compare our method to (Ng, Jordan, and Weiss, 2002). Results for spectral clustering are shown in Fig. 2. Note that for the same clustering problems, SC performs poorly, unlike our method which finds the *right* clusters. We should remark, however, that the concept of *right* clustering has not been defined, our solutions simply seem to be much more consistent with answers from human observers.

More formally, by not assuming that the *correct* similarity measure is known or that it is the same for all classes, our algorithm is able to perform well in a larger class of problems than other methods. Our algorithm also discovers the intrinsic correct measure that would generate clusters with higher likelihood. For example in cases when both clusters share the same inner scale, both SC and our method perform well, and our method discover that the inner scales are the same for both classes, as seen in Fig. 4(a)-(b). However, when the right clustering configurations requires that each class be treated different in terms of the inner similarity measure, our method clearly outperforms SC, and is able to infer the class dependent measures that are adequate for the correct clustering. It is important to note that neighboring points in the same class are not necessarily separated by the same distance. The clusters may have non-convex complex shapes or even overlap onto each other as in Fig. 4(c)-(i) so it may seen there is no real way to evaluate their correctness. However, they seem to agree with human judgment (recall that our problem is completely unsupervised). Certain connections have been established between unsupervised clustering and Gestalt theory (Perona and Freeman, 1998; Ng, Jordan, and Weiss, 2002). Our clustering criterion is well suited for problems where the point density is not uniform across classes and there is no clear separation between these classes. For example, problems where missing data occurs at different rates depending on the class.

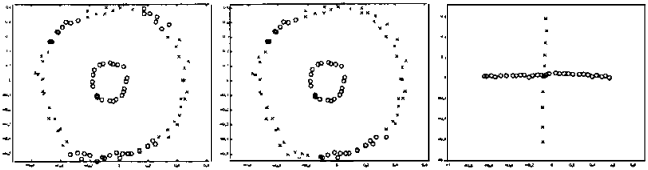

Figure 5: Examples in left-center use different K-neighbor prior. Example in right uses a exponential likelihood. See Fig. 4 for explanation.

We could also use the neighborhood prior in Eq. 9. In this case, we need to set one parameter, the number of neighbors $K$. We can see in Fig. 5(left-center) that for some data sets where the connected graph prior performs well, this prior cannot find the correct clustering (we searched over many $K$'s for a good number of neighbors). We observed similar behavior for the rest of the datasets also. An exponential likelihood (Eq. 8) can also be used, and a similar inference theorem can be derived. Although we do not prove it here, Fig. 5(right) shows an example solution. Results for this likelihood are similar compared to the Gaussian likelihood. This may seem a more suitable likelihood for our specific definition of $L$ since elements in $L_{ij}$ are positive. It is also more related to the measure used in SC.

## 5 DISCUSSION

We have developed and analyzed two new classes of clustering models. Using these models, we have established a connection between the widely used SC algorithm and probabilistic models. This may open the door to new approximate methods for SC, for example based on inference in probabilistic models, and may also provide much desired tools for analysis. This paper also shows how one can infer the similarity measure for improving clustering performance in problems where the clusters overlap or the correct clustering implies having a different inner measure for the different clusters. These problems cannot be consistently solved by methods such as spectral, mixture, K-means, and connected components clustering. In this class of problems, inferring $C$ is computationally unfeasible even for small data sets. Despite this, we have developed a technique for optimizing for $C$ efficiently.

The notions introduced here have potentially interesting extensions. For example, this method has implications to semi-supervised clustering (Szummer and Jaakkola, 2002; Xing et al., 2003). A simple extension would just involve assigning the known labels to the corresponding random variables $c_i$ and running the algorithm as usual. Our method could use, but does not require, this form of side-information. It is unlikely that any distance metric of the form proposed in (Xing et al., 2003); or a random walk approach, with equal similarity measures for all classes, could produce the solutions shown in our examples.

## Acknowledgements

We thank T. Jaakkola for motivating discussions. We appreciate the reviewers feedback and clarifying questions.